# The Logic for a Mildly Context-Sensitive Fragment of the Lambek-Grishin Calculus

Hiroyoshi Komatsu

Keio University, Japan
hkomatsu@nak.ics.keio.ac.jp

**Abstract.** While context-free grammars are characterized by a simple proof-theoretic grammatical formalism namely categorial grammar and its logic the Lambek calculus, no such characterizations were known for tree-adjoining grammars, and even for any mildly context-sensitive languages classes in the last forty years despite some efforts [15, 13]. We settle this problem in this paper. On the basis of the existing fragment of the Lambek-Grishin calculus which captures tree-adjoining languages, we present a logic called **HLG**: a proof-theoretic characterization of tree-adjoining languages based on the Lambek-Grishin calculus [9, 10] restricted to Hyperedge-replacement grammar with rank two studied by Moot [13, 14]. **HLG** is defined in display calculus with cut-admissibility. Several new techniques are introduced for the proofs, such as purely structural connectives, usefulness, and a graph-theoretic argument on proof nets for **HLG**.

## 1 Introduction

The Lambek-Grishin calculus (**LG**) [4] is a logic originally defined by Goré [3] and rediscovered for a linguistic application by Moortgat [9, 10]. This logic is based on algebraic generalizations of the Non-associative Lambek calculus (**NL**) studied by Grihin [4]. **LG** is defined in display calculus with cut-admissibility and can capture several language phenomena that cannot be captured in the Non-associative Lambek calculus (**NL**) and equivalently context-free grammars. Unlike **NL**, however, the complexity of **LG** is known as NP-complete [1]. This implies a negative result: parsing a natural language sentence in **LG** is not done in a realistic time.

On the other hand, Moot showed that the Lambek-Grishin calculus contains a "fragment" that can be transformed into tree-adjoining grammars by using hyperedge-replacement grammars [13, 14]. This fragment contains all the known linguistic examples for the Lambek-Grishin calculus shown in the original paper [10]. However, since this fragment is characterized by the property, namely planarity in proof nets for **LG**, it does not correspond to a specific logic defined as a sequent calculus or generalizations of it.

This paper resolve this gap by showing that the mildly context-sensitive fragment of **LG** [13] can be axiomatized as a logic which enjoys cut-admissibility on display calculi. We prove that the proof-search problem of this logic is in PTIME.

We call the logic HR$_2$-restricted LG (**HLG**). As for linguistic application, **HLG** can capture the mildly context-sensitive language phenomena illustrated in the papers for **LG** [9, 10], including the phenomena called infixation, extraction and cross-serial dependencies.

It is known that context-free grammars are characterized by the non-associative Lambek calculus as categorial grammars studied in 1950s [7]. On the other hand, tree-adjoining grammars [5] and other mildly context-sensitive grammars have no such a characterization as a sequent calculus or a generalization of it. As a characterization of tree-adjoining languages, combinatory categorial grammars (CCG) [15] are one of existing approaches. However, since the rules in CCG break ordinary requirements for logics defined in sequent calculi or generalizations of them, it is no longer able to be contextualized among the studies of other logics directly. For example, the type-raising rule $X \to T/(X\backslash T)$ in CCG[1] is appliable if $T$ is an atomic type (not appliable for formulae in general). This breaks familiar conditions required to define a logic in proof theory. On the other hand, **HLG** is defined as a display calculus, a generalization of sequent calculi, and indeed similar arguments with other substructural logics are exploited in this paper.

In §2, we describe the Lambek-Grishin calculus and the Lambek system, and proof nets. In §3, we present **HLG** and analyze its expressibility, linguistic application, and proof nets for **HLG** used for the proof of its expressibility. it is defined on display calculi, but a natural extension of display calculi called purely structural connectives is used. In addition, the property called usefulness, an analogy of (cut-)admissibility for typelogical grammars, is used for simplifying the axiomatization of **HLG**.

## 2  Preliminary

In this section, we review the basic materials used in this paper, the Lambek-Grishin calculus [9, 10] and its proof nets [11].

### 2.1  Lambek-Grishin calculus

The Lambek-Grishin calculus is defined on Belnap's display calculi [3], a generalization of Gentzen sequent calculus. The notation given here is inherited from Moortgat's [10] who gives the name of the Lambek-Grishin calculus. However, the definition of it on the display calculi can be traced in the origin [3] motivated by Grishin algebras [4].

The Lambek-Grishin calculus (**LG**) is a generalization of the Non-associative Lambek calculus (**NL**) with symmetric connectives. **NL** is defined on the connectives, product, left and right division ($\otimes, /, \backslash$), here we call it the Lambek's connectives, and **LG** is extended to the symmetric connectives, coproduct, right and left division ($\oplus, \oslash, \obslash$) called Grishin's connectives. The interaction between

---

[1] We represent the rule in the original Lambek's notation rather than the CCG notation

Lambek's and Grishin's are given by "linear distributivity" or called Grishin's interactions. In the original paper by Grishin [4], there are four types of interactions, but now Grishin's interaction mostly refers to his fourth interaction, Grishin's interaction IV. See also [10].

In display calculi, the rules are defined on the structures rather than formulae. This permits us to define a rule that breaks cut-admissibility on the usual Gentzen sequent calculus. See also [3] for the comprehensive view in the context of structural logics. We denote structural connectives with black dots. For example, product on the formulae is denoted by $(A \otimes B)$, and on the structures denoted by $(A \cdot \otimes \cdot B)$.

The formula of **LG** is defined as follows:

$$\begin{aligned} A, B \quad ::= \quad & \alpha \mid \\ & A \otimes B \mid B \backslash A \mid A / B \mid \\ & A \oplus B \mid A \oslash B \mid B \obslash A \end{aligned}$$

In a sequent $\Gamma \vdash \Theta$, its antecedent (succedent) is $\Gamma$ ($\Theta$). The antecedent (succedent) in display calculi must be an element of the set called the **input structure** (**output structure**). Let $\mathcal{F}$ be a formula. The input structure $\mathcal{I}$ and output strcture $\mathcal{O}$ are defined as follows:

$$\begin{aligned} \mathcal{I} \quad ::= \quad & \mathcal{F} \mid \mathcal{I} \cdot \otimes \cdot \mathcal{I} \mid \mathcal{I} \cdot \obslash \cdot \mathcal{O} \mid \mathcal{O} \cdot \oslash \cdot \mathcal{I} \\ \mathcal{O} \quad ::= \quad & \mathcal{F} \mid \mathcal{O} \cdot \oplus \cdot \mathcal{O} \mid \mathcal{I} \cdot \backslash \cdot \mathcal{O} \mid \mathcal{O} \cdot / \cdot \mathcal{I} \end{aligned}$$

We present the axiomatization of **LG** in the display calculi. In display calculi, to control structural levels, there are two types of rules. The rules denoted by one lines is defined as directional, which transfers an upper sequent to a lower sequent, the rules denoted by double lines are bidirectional rules, an upper sequent to a lower sequent and vice versa is allowed.

The display calculus for **LG** is defined as follows:

$$\frac{}{A \vdash A} \, ax \qquad \frac{X \vdash A \quad A \vdash Y}{X \vdash Y} \, cut$$

*Monotonicity principles*

$$\frac{A \vdash B \quad C \vdash D}{A \cdot \otimes \cdot C \vdash B \otimes D;} \, R\otimes, \, L/, \, L\backslash \qquad \frac{A \vdash B \quad C \vdash D}{A \oplus C \vdash B \cdot \oplus \cdot D;} \, L\oplus, \, R\oslash, \, R\obslash$$
$$A/D \vdash B \cdot / \cdot C; \qquad\qquad\qquad A \cdot \oslash \cdot D \vdash B \oslash C;$$
$$D \backslash A \vdash C \cdot \backslash \cdot B \qquad\qquad\qquad D \cdot \obslash \cdot A \vdash C \obslash B$$

*(Dual) residuation principles*

$$\frac{\dfrac{B \vdash A \cdot \backslash \cdot C}{A \cdot \otimes \cdot B \vdash C}}{A \vdash C \cdot / \cdot B} \begin{array}{l} rp \\ rp \end{array} \qquad \frac{\dfrac{C \cdot \oslash \cdot A \vdash B}{C \vdash B \cdot \oplus \cdot A}}{B \cdot \obslash \cdot C \vdash A} \begin{array}{l} drp \\ drp \end{array}$$

*Grishin distributivity principles (IV)*

$$\frac{A \cdot \otimes \cdot B \vdash C \cdot \oplus \cdot D}{C \cdot \oslash \cdot A \vdash D \cdot / \cdot B} G_1 \qquad \frac{A \circ \otimes \circ B \vdash C \cdot \oplus \cdot D}{C \cdot \oslash \cdot B \vdash A \circ \backslash \circ D} G_2$$

$$\frac{A \cdot \otimes \cdot B \vdash C \cdot \oplus \cdot D}{B \cdot \oslash \cdot D \vdash A \cdot \backslash \cdot C} G_3 \qquad \frac{A \circ \otimes \circ B \vdash C \cdot \oplus \cdot D}{A \cdot \oslash \cdot D \vdash C \circ / \circ B} G_4$$

*Rewrite principles*

$$\frac{A \cdot \otimes \cdot B \vdash Y}{A \otimes B \vdash Y} L\otimes \qquad \frac{A \cdot \oslash \cdot B \vdash Y}{A \oslash B \vdash Y} L\oslash \qquad \frac{A \cdot \oslash \cdot B \vdash Y}{A \oslash B \vdash Y} L\oslash$$

$$\frac{X \vdash A \cdot \oplus \cdot B}{X \vdash A \oplus B} R\oplus \qquad \frac{X \vdash A \cdot / \cdot B}{X \vdash A/B} R/ \qquad \frac{X \vdash A \cdot \backslash \cdot B}{X \vdash A \backslash B} R\backslash$$

### 2.2 Lambek system

The Lambek system is an interpretation of a logic as a grammatical formalism. Consider a grammar with a logic defined as a tuple $(\alpha, \Sigma, L, s)$ where $\alpha$ is atoms, $\Sigma$ is alphabet such that $\alpha \cap \Sigma = \emptyset$, $L$ is a lexical relation such that $L \subseteq \Sigma \times F$, where $F$ is defined as formula corresponds to $\alpha$, $s$ is a distinguished symbol such that $s \in \alpha$. Let $\bigotimes_{i=1}^{n} w_i$ be a formula that is one of possible products of $t_1, ..., t_n$ in order where $t_i \in L(w_i)$ for $i \in [n]$. (Recall that the product is non-associative.) The string $S = w_1, ..., w_n$ is said to be a **sentence** if a sequent $\bigotimes_{i=1}^{n} w_i \vdash s$ is provable in **LG**. Let $\mathcal{L}$ be the set of all languages recognizable in **LG**.

### 2.3 Proof Nets

Proof nets are originally introduced as a geometrical representation and a normal form of proofs in the context of linear logic [2]. On the later Moot extended the idea for display calculi [12]. Proof nets for display calculi normalize the differences of proofs arisen from the (dual) residuation principles. See [11] for the definition of proof nets for **LG**.

## 3   HR$_2$-restricted LG

In this section, we give an axiomatization of HR$_2$-restricted LG (**HLG**) in display calculi with some linguistic example that illustrates language phenomena so-called displacement that are problematic for the Lambek calculus, and prove that the generative capacity of **HLG** on the Lambek system is indeed tree-adjoining languages.

The key idea of **HLG** is axiomatization of Moot's fragment [13] using a *purely* structural control of connectives – denoted by the symbols with white dots (∘). While the structural connectives of display calculi are still allowed to deduce the corresponding formula, in **HLG** the purely structural connectives are not allowed to be "weakened" to the corresponding formula from it. For example in display calculi, recall that the structural connective of product is denoted by $(A \cdot \otimes \cdot B)$ in **LG**, the purely structural connective of it is denoted by $(A \circ \otimes \circ B)$, and we prohibit to weaken it to $(A \otimes B)$. This new type of logics allows more strong structural controls than other display calculi. The deduced sequent used in the Lambek system is therefore defined as a pair of structures rather than a pair of formulae defined in §2.2.

We begin to define **HLG**. The formula of **HLG** is defined similarly to **LG** as follows:

$$A, B ::= \alpha \mid$$
$$A \otimes B \mid B \backslash A \mid A / B \mid$$
$$A \oplus B \mid A \oslash B \mid B \obslash A$$

The input structure and the ouput structure are recursively defined as follows where $\mathcal{F}$ stands for the set of the all formulae in **HLG**.

$$\mathcal{I} ::= \mathcal{F} \mid \mathcal{I} \cdot \otimes \cdot \mathcal{I} \mid \mathcal{I} \cdot \oslash \cdot \mathcal{O} \mid \mathcal{O} \cdot \oslash \cdot \mathcal{I} \mid \mathcal{I} \circ \otimes \circ \mathcal{I} \mid \mathcal{I} \circ \oslash \circ \mathcal{O} \mid \mathcal{O} \circ \oslash \circ \mathcal{I}$$
$$\mathcal{O} ::= \mathcal{F} \mid \mathcal{O} \cdot \oplus \cdot \mathcal{O} \mid \mathcal{I} \cdot \backslash \cdot \mathcal{O} \mid \mathcal{O} \cdot / \cdot \mathcal{I} \mid \mathcal{O} \circ \oplus \circ \mathcal{O} \mid \mathcal{I} \circ \backslash \circ \mathcal{O} \mid \mathcal{O} \circ / \circ \mathcal{I}$$

We call the logical system defined by the following rules on display logic **HLG**:

$$\frac{}{A \vdash A} \ ax \qquad \frac{X \vdash A \quad A \vdash Y}{X \vdash Y} \ cut$$

*Monotonicity principles*

$$\frac{A \vdash B \quad C \vdash D}{A \cdot \otimes \cdot C \vdash B \otimes D;} \ R\otimes, \ L/, \ L\backslash \qquad \frac{A \vdash B \quad C \vdash D}{A \oplus C \vdash B \cdot \oplus \cdot D;} \ L\oplus, \ R\oslash, \ R\obslash$$
$$A/D \vdash B \cdot / \cdot C; \qquad\qquad\qquad A \cdot \oslash \cdot D \vdash B \oslash C;$$
$$D\backslash A \vdash C \cdot \backslash \cdot B \qquad\qquad\qquad D \cdot \obslash \cdot A \vdash C \obslash B$$

*(Dual) residuation principles*

$$\frac{B \vdash A \cdot \backslash \cdot C}{\frac{A \cdot \otimes \cdot B \vdash C}{A \vdash C \cdot / \cdot B} \ rp} \ rp \qquad \frac{C \cdot \oslash \cdot A \vdash B}{\frac{C \vdash B \cdot \oplus \cdot A}{B \cdot \obslash \cdot C \vdash A} \ drp} \ drp$$

*Weak (dual) residuation principles*

$$\frac{B \vdash A \circ \backslash \circ C}{\frac{A \circ \otimes \circ B \vdash C}{A \vdash C \circ / \circ B} \ rp} \ rp \qquad \frac{C \circ \oslash \circ A \vdash B}{\frac{C \vdash B \circ \oplus \circ A}{B \circ \obslash \circ C \vdash A} \ drp} \ drp$$

*structural weakening principles* $(\% \in \{\oplus, /, \backslash\}, \# \in \{\otimes, \oslash, \obslash\})$

$$\frac{A \cdot \% \cdot B \vdash Y}{A \circ \% \circ B \vdash Y} \ w\% \qquad \frac{X \vdash A \cdot \# \cdot B}{X \vdash A \circ \# \circ B} \ w\#$$

*Grishin linear distributivity principles (IV)*

$$\frac{A \cdot \otimes \cdot B \vdash C \cdot \oplus \cdot D}{C \cdot \oslash \cdot A \vdash D \cdot / \cdot B} \; G_1 \qquad \frac{A \circ \otimes \circ B \vdash C \cdot \oplus \cdot D}{C \cdot \oslash \cdot B \vdash A \circ \backslash \circ D} \; G_2$$

$$\frac{A \cdot \otimes \cdot B \vdash C \cdot \oplus \cdot D}{B \cdot \oslash \cdot D \vdash A \cdot \backslash \cdot C} \; G_3 \qquad \frac{A \circ \otimes \circ B \vdash C \cdot \oplus \cdot D}{A \cdot \oslash \cdot D \vdash C \circ / \circ B} \; G_4$$

*Rewrite principles*

$$\frac{A \cdot \otimes \cdot B \vdash Y}{A \otimes B \vdash Y} \; L\otimes \qquad \frac{A \cdot \oslash \cdot B \vdash Y}{A \oslash B \vdash Y} \; L\oslash \qquad \frac{A \cdot \oslash \cdot B \vdash Y}{A \oslash B \vdash Y} \; L\oslash$$

$$\frac{X \vdash A \cdot \oplus \cdot B}{X \vdash A \oplus B} \; R\oplus \qquad \frac{X \vdash A \cdot / \cdot B}{X \vdash A/B} \; R/ \qquad \frac{X \vdash A \cdot \backslash \cdot B}{X \vdash A \backslash B} \; R\backslash$$

Similarly to **LG**, we call by **HLG**$_\emptyset$ the logic **LG** without Grishin linear distributivity principles IV.

The important difference between **HLG** and **LG** is the purely structural connectives used in the rules $(G_2), (G_4)$. The residuation rules and structural weakening principles are defined to be followed from axiomatization using purely structural connectives. On the later we can easily see that **HLG** is a fragment of **LG** in terms of the weak generative capacity on the Lambek system. The proof in Appendix A shows that for every proof $P$ in **LG**, the proof net of $P$ satisfies with *planarity* on the graph if and only if $P$ can be proved in the system **HLG**, and therefore it satisfies with the criteria of Moot's fragment of **LG**.

### 3.1 Lambek system for HLG

We define the Lambek system for **HLG**. Let $s$ be the distinguished symbol that indicates a valid sentence. We define the lexical relation $L(w)$, mapping from a word $w$ to the finite set of lexical formulae in the same way as **LG** and **NL**. The language $L$ is said to be in **HLG** if, for every string $(s_1, s_2, ..., s_n) \in L$, there exists a word $w_i \in L(s_i)$ such that one of the sequents

$$w_1 \circ \otimes \circ ... \circ \otimes \circ w_n \vdash s$$

is provable in **HLG**.

### 3.2 Linguistic application

The cross-serial dependency, a well-known counterexample in linguistics, is phenomenon cannot be captured in the context-free grammars but can be in tree-adjoining grammars [5] (See more details in [6] for example). We show that this phenomenon can be indeed captured by **HLG**. For example, let us consider the following cross-serial dependency sentences in Dutch:

(1)    (omdat) ik Cecilia de nijlpaarden zag voeren.

(2)    (omdat) ik Cecilia Henk de nijlpaarden zag helpen voeren.

The sentence 2 has the dependency between the words "Henk" and "helpen". Note that the other words in 2 has the same dependencies as 1. In the following we abbreviate "de nijlpaarden" by "dn".

The lexical map is defined as follows where we distinguished $vp$ and $vp'$ only for clarification:

$$\text{zag} ::= (vp \oslash (np\backslash(np\backslash s))) \oslash vp'$$
$$\text{helpen, voeren} ::= vp'\backslash(np\backslash vp)$$
$$\text{ik, Cecilia, Henk, dn} ::= np$$

The sentence 1 is in **HLG** as follows:

$$
\cfrac{
\cfrac{
\cfrac{
\cfrac{
\cfrac{
\cfrac{
\cfrac{\vdots}{np \cdot \otimes \cdot (vp' \cdot \otimes \cdot (vp'\backslash(np\backslash vp))) \vdash (vp \oslash (np\backslash(np\backslash s))) \cdot \oplus \cdot (np \cdot \backslash \cdot (np \cdot \backslash \cdot s))}
}{np \circ \otimes \circ (vp' \cdot \otimes \cdot (vp'\backslash(np\backslash vp))) \vdash (vp \oslash (np\backslash(np\backslash s))) \cdot \oplus \cdot (np \cdot \backslash \cdot (np \cdot \backslash \cdot s))} w\otimes
}{(vp \oslash (np\backslash(np\backslash s))) \cdot \oslash \cdot (vp' \cdot \otimes \cdot (vp'\backslash(np\backslash vp))) \vdash np \circ \backslash \circ (np \cdot \backslash \cdot (np \cdot \backslash \cdot s))} G_2
}{vp' \cdot \otimes \cdot (vp'\backslash(np\backslash vp)) \vdash (vp \oslash (np\backslash(np\backslash s))) \cdot \oplus \cdot (np \circ \backslash \circ (np \cdot \backslash \cdot (np \cdot \backslash \cdot s)))} drp
}{(vp \oslash (np\backslash(np\backslash s))) \cdot \oslash \cdot vp' \vdash (np \circ \backslash \circ (np \cdot \backslash \cdot (np \cdot \backslash \cdot s))) \cdot / \cdot (vp'\backslash(np\backslash vp))} G_1
}{(vp \oslash (np\backslash(np\backslash s))) \oslash vp' \vdash (np \circ \backslash \circ (np \cdot \backslash \cdot (np \cdot \backslash \cdot s))) \cdot / \cdot (vp'\backslash(np\backslash vp))} L\oslash
}{\underbrace{np}_{\text{ik}} \circ \otimes \circ \underbrace{np}_{\text{Cecilia}} \circ \otimes \circ \underbrace{np}_{\text{dn}} \circ \otimes \circ \underbrace{(vp \oslash (np\backslash(np\backslash s))) \oslash vp'}_{\text{zag}} \circ \otimes \circ \underbrace{(vp'\backslash(np\backslash vp))}_{\text{voeren}} \vdash s} rp, w\otimes
$$

We can deal with the sentences 1 and 2 by merging the two symbols $vp$ and $vp'$ into one symbol.

### 3.3 HLG$_\bullet$

**HLG$_\bullet$** is an extension of **HLG** to prove the correspondence between proofs in **LG** whose proof nets are planar and **HLG** shown in Appendix A. Even though the definition of **HLG$_\bullet$** is more complicated than **HLG**, as for generative capacity we prove that **HLG$_\bullet$** and **HLG** are equivalent. Therefore, **HLG$_\bullet$** is an intermediate representation to show our main theorem in a nutshell.

We call by **HLG$_\bullet$** the logic **HLG$_\emptyset$** with following axioms:

*Grishin linear distributivity principles (IV) for planarity*

$$\frac{A \cdot \otimes \cdot B \vdash C \cdot \oplus \cdot D}{C \cdot \oslash \cdot A \vdash D \cdot / \cdot B} G_1$$

$$\frac{A \cdot \otimes \cdot B \vdash C \circ \oplus \circ D}{C \cdot \oslash \cdot A \vdash D \circ / \circ B} G_1 L \qquad \frac{A \circ \otimes \circ B \vdash C \cdot \oplus \cdot D}{C \circ \oslash \circ A \vdash D \cdot / \cdot B} G_1 R$$

$$\frac{A \circ \otimes \circ B \vdash C \cdot \oplus \cdot D}{C \cdot \oslash \cdot B \vdash A \circ \backslash \circ D} G_2 L \qquad \frac{A \cdot \otimes \cdot B \vdash C \circ \oplus \circ D}{C \circ \oslash \circ B \vdash A \cdot \backslash \cdot D} G_2 R$$

$$\frac{A \cdot \otimes \cdot B \vdash C \cdot \oplus \cdot D}{B \cdot \oslash \cdot D \vdash A \cdot \backslash \cdot C} \; G_3$$

$$\frac{A \cdot \otimes \cdot B \vdash C \circ \oplus \circ D}{B \cdot \oslash \cdot D \vdash A \circ \backslash \circ C} \; G_3 L \qquad \frac{A \circ \otimes \circ B \vdash C \cdot \oplus \cdot D}{B \circ \oslash \circ D \vdash A \cdot \backslash \cdot C} \; G_3 R$$

$$\frac{A \circ \otimes \circ B \vdash C \cdot \oplus \cdot D}{A \cdot \oslash \cdot D \vdash C \circ / \circ B} \; G_4 L \qquad \frac{A \cdot \otimes \cdot B \vdash C \circ \oplus \circ D}{A \circ \oslash \circ D \vdash C \cdot / \cdot B} \; G_4 R$$

These rules are designed to preserve planarity in proof nets for $\mathbf{HLG_\bullet}$. Notice that the logic $\mathbf{HLG_\emptyset}$ with $(G_1, G_2L, G_3, G_4L)$ rules is completely the same definition as $\mathbf{HLG}$. From now on, we often ignore the differences of names between $(G_2), (G_4)$ and $(G_2L), (G_4L)$ respectively for the compatibility if there is no ambiguity.

For the equivalence between $\mathbf{HLG}$ and $\mathbf{HLG_\bullet}$ on the Lambek system, we are going to introduce the property called usefulness and give proofs by a *mutual* induction on proofs in $\mathbf{HLG_\bullet}$ in §3.6.

### 3.4   Planarity in proof nets

We show an example formula whose proof net in $\mathbf{LG}$ is not planar, and thus is provable in $\mathbf{LG}$ but not in $\mathbf{HLG}$.

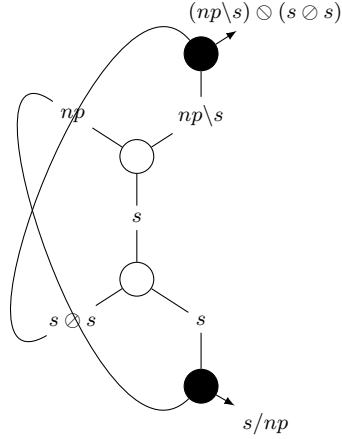

**Fig. 1.** The abstract proof structure for the sequent $(np\backslash s) \oslash (s \oslash s) \vdash s/np$ breaks the condition of planarity in the proof net.

Figure 1 shows that the proof net of the sequent $(np\backslash s) \oslash (s \oslash s) \vdash s/np$ does not satisfy with planarity. The proof of this sequent is provable in $\mathbf{LG}$ as

follows:

$$\cfrac{\cfrac{\vdots}{\cfrac{np \cdot \otimes \cdot np\backslash s \vdash (s \oslash s) \cdot \oplus \cdot s}{\cfrac{(s \oslash s) \cdot \odot \cdot (np\backslash s) \vdash np \cdot \backslash \cdot s}{(s \oslash s) \odot (np\backslash s) \vdash np\backslash s}\ L\odot, R\backslash}}}{}\ G_2$$

On the other hand, in **HLG**$_\bullet$ and **HLG**, the both rules ($L\odot$) and ($R\backslash$) are incompatible as follows:

$$\cfrac{\cfrac{\cfrac{\vdots}{\cfrac{np \cdot \otimes \cdot np\backslash s \vdash (s \oslash s) \cdot \oplus \cdot s}{np \circ \otimes \circ np\backslash s \vdash (s \oslash s) \cdot \oplus \cdot s}\ w\otimes}}{\cfrac{(s \oslash s) \cdot \odot \cdot (np\backslash s) \vdash np \circ \backslash \circ s}{(s \oslash s) \odot (np\backslash s) \vdash np \circ \backslash \circ s}\ L\odot}}{}\ G_2L \qquad \cfrac{\cfrac{\cfrac{\vdots}{\cfrac{np \cdot \otimes \cdot np\backslash s \vdash (s \oslash s) \cdot \oplus \cdot s}{np \cdot \otimes \cdot np\backslash s \vdash (s \oslash s) \circ \oplus \circ s}\ w\oplus}}{\cfrac{(s \oslash s) \circ \odot \circ (np\backslash s) \vdash np \cdot \backslash \cdot s}{(s \oslash s) \circ \odot \circ (np\backslash s) \vdash np\backslash s}\ R\backslash}}{}\ G_2R$$

### 3.5 Proof nets

The proof nets for **HLG** and **HLG**$_\bullet$ are defined similarly to **LG**. The only difference is the proof structure for the purely structural connectives. We call the links for the new structural connectives $X \circ \% \circ Y$ ($\% \in \{\oplus, /, \backslash, \otimes, \oslash, \odot\}$) the purely tensor links. For example, the proof structure of $A/B \circ \otimes \circ B \vdash A$ denoted by a purely tensor link is as follows:

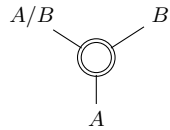

To prove **HLG** has only planar proof nets in Appendix A, we exploit the one-way transformation property of links: the tensor links can be transformed to the purely tensor links using the structural weakening principles, but the purely tensor links cannot be transformed to the tensor links.

For purely structural connectives, some of the structural rules in **HLG** and **HLG**$_\bullet$ are modified using the representation of purely tensor links, the transfor-mation of abstract proof structures by $(G_1L)$, $(G_3L)$, $(G_1R)$, $(G_3R)$ are shown in Fig 2-3.

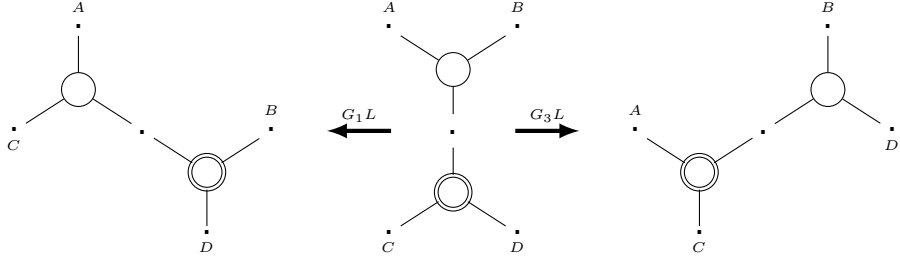

**Fig. 2.** $(G_1L)$ and $(G_3L)$ — "mixed associativity I"

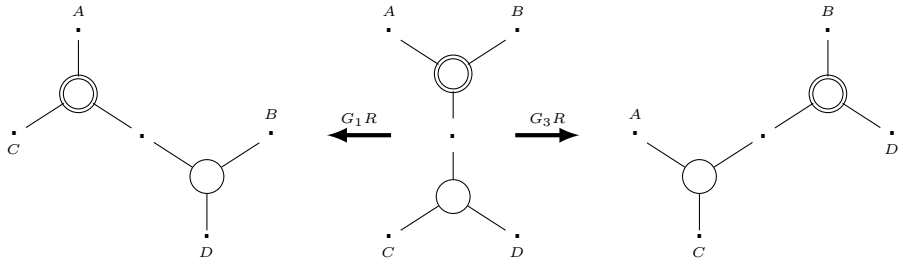

**Fig. 3.** $(G_1R)$ and $(G_3R)$ — "mixed associativity II"

Similarly, the transformations of $(G_4L)$, $(G_2L)$, $(G_4R)$, $(G_2R)$ are shown in Fig. 4-5.

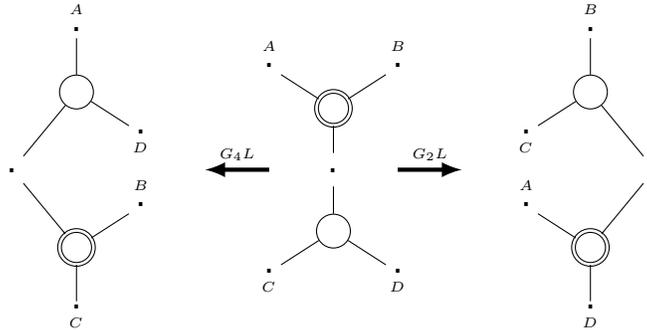

**Fig. 4.** $(G_4L)$ and $(G_2L)$ — "mixed purely commutativity I"

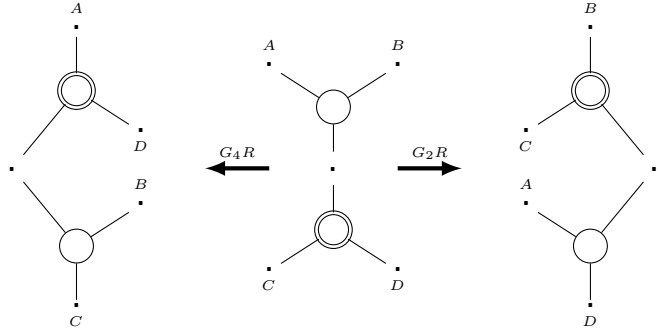

**Fig. 5.** $(G_4R)$ and $(G_2R)$ — "mixed purely commutativity II"

The important property of proof nets for **HLG** is planarity. The following theorem holds.

**Theorem 1.** *Every proof $P$ in **LG** has a planar proof nets if and only if $P$ is provable in **HLG**.*

*Proof.* By a graph-theoretic analysis of proof nets. The proof is given in the appendix A.

### 3.6 Usefulness

**Notation** A (Grishin) rule $r$ is said to be **useless** if the logical system without the rule $r$ has the same language class $\mathcal{L}$ on the Lambek system. Note that this is the analogous to the definition of *admissible* rules, such as cut-admissibility, for the Lambek systems. Let us call the sequence of applying structural rules a *deduction*. A proof $P$ is said to be proved using the deduction $R$ if one can prove $P$ by applying the structural rules only allowed to apply as a sequence $s \in \mathcal{G}$ in $P$. The set of deductions $\mathcal{G}$ is said to be *useless* if for every sentence that can be proved using $\mathcal{G}$ can also be proved in $\mathbf{HLG}_\emptyset$.

**Generative capacity of $\mathbf{HLG}_\bullet$** The set of all the languages generated in $\mathbf{HLG}_\bullet$ exactly corresponds to the Tree-Adjoining Languages shown in Appendix A, and one of the useful sets of deductions is the combination of $(G_2)$ and $(G_1)$. In other words, the set of deductions $\mathcal{G}_{\{(G_2),(G_1)\}}$ defined as the set of sequences $(\nabla_i)_{i=1}^n$ for every $\nabla_i \in \{(G_2),(G_1)\}$ is useful (not useless). Therefore it is proved that each of the six rules of Grishin linear distributivity principles (I V) for planarity is useless. However, we prove this fact from a syntactic argument.

**Theorem.** *Each of the rule $(G_1R)$ and $(G_3R)$ is useless.*

*Proof.* An argument by a variant of induction. Let $m$ be the purely multiplicative structure of the assumption in the upper sequent of $(G_1R)$. For every application of $(G_1R)$ in a proof $P$ of a sentence, the rule that operated the purely multiplicative $m$ can be classified as follows:

(a) $(w\otimes)$,

(b) $(G_1L)$ or $(G_3L)$,

(c) $(G_2)$ or $(G_4)$.

In the case (a) the application of $(G_1R)$ can be replaced by $(G_1)$. In the case (c) it is possible to repeatedly apply $(G_2)$ or $(G_4)$ on $m$, but the deductions are never of form a $n$-ary relation on the words ($n \geq 3$). Let $m_v$ be resulting purely multiplicative structure of the two formulae $A, B$. The formulae $A, B$ can be replaced by other two formulae because every binary relation of the consecutive two phrases $A, B$ (More preciously, the purely multiplicative of two formulae) can be represented by the connectives $(/, \backslash)$ in the modified lexical relation. In this case the purely multiplicative $m_v$ is still remained, so it can be reduced to another case without changing the relation of the words. In the case (b), the conclusion of the upper sequent of $(G_1L)$ or $(G_3L)$ is a dual purely multiplicative. Therefore, we can classify the rule that operates the purely multiplicative in $P$ in a similar way:

(I) $(w\oplus)$,

(II) $(G_1R)$ or $(G_3R)$,

(III) $(G_2R)$ or $(G_4R)$.

In the cases (I) and (III), we can apply the same argument above. In the case (II) we apply mutual recursion with the classifications (a)-(c) and (I)-(III), and obtain the final case that falls into the case (a) or (I). This recursive case yields a $n$-ary relation of words where $n$ is the number of the recursion depth, but the application of these rules can be replaced by the repetition of $(G_1)$ or $(G_3)$ to construct the same relation since $(w\otimes)$ and $(w\oplus)$ can be applied at the last instead. The case of $(G_3R)$ is proved in the same way. As the later part of the proof shows, the following theorems hold:

**Corollary.** *Each of $(G_1L)$ and $(G_3L)$ is useless.*

**Corollary.** *Each of $(G_2R)$ and $(G_4R)$ is useless.*

*Proof.* The rule $(G_2R)$ has a dual purely structural connective in the lower sequent. Since now there is no rules to eliminate it, $(G_2R)$ is not used in any proof of a sentence. The case of $(G_4R)$ is proved by the same argument.

From these theorem and corollaries, the sets of languages of **HLG** and **HLG**$_\bullet$ are equivalent.

### 3.7  Cut-admissibility

The proof of the cut-elimination theorem for **LG**$_{\mathbf{I+IV}}$ shown in [10] still works for **HLG**. Consider the cut rule shown below. Since $A$ is a formula, $A$ does not contain any purely structural connectives $X \circ \% \circ Y$ ($\% \in \{\oplus, /, \backslash, \otimes, \oslash, \otimes\}$), and since the rules in **HLG** without any mention of the purely structural connectives are completely same rules as in **LG**, we can prove it using the same induction of the length of $A$ shown in [10].

$$\frac{X \vdash A \quad A \vdash Y}{X \vdash Y} \ cut$$

## A  Expressibility of HLG

We prove Theorem 1 that states every proof $P$ of **LG** has a planar proof nets if and only if $P$ is provable in **HLG**. By §3.6, it is sufficient to consider **HLG**$_\bullet$ instead of **HLG**.

To prove Theorem 1, we use some definitions and lemmas. For simplicity, an abstract proof net is called a **P-graph**. We start from some notations.

**Definition 1.** *A **sub P-graph** of a P-graph A is a graph formed from a subset of the vertices and hyperedges of A. The structural rules $(G_1)$, $(G_2)$, $(G_3)$, $(G_4)$ are called **Grishin rules**. The P-graph before applying the Grishin rules on Fig 8 and Fig 9 is called **basis** (or root and branch). The bunch of two links of the assumption on bases are called **branch**. The bunch of two links of the conclusion on bases are called **root**. Applying the Grishin rules to a basis is said to be a transformation (of the basis). In addition, it is said to be **growing a basis** (by a transformation) if the new basis appears in the graph by the transformation. The rules $(R/)$, $(L\otimes)$, $(R\backslash)$, $(L\oslash)$, $(R\oplus)$, $(L\obackslash)$ are called **discharging rules**. If a cotensor link $l_{co}$ is disappeared from the graph by a discharging rule, it is said that $l_{co}$ **is discharged**. The **knot** (of two links) in a (sub) P-Graph A is defined as the crossing vertexes of the two links.*

In the following, we assume that the Grishin rules are applied after applying a discharging rule if possible to identify abstract proof nets corresponds to a formula.

**Definition 2.** *For every P-graph A, we denote by $\Gamma \Rightarrow \Delta$ a basis $\Gamma$ in A is transformed to a sub P-graph $\Delta$. For every P-graph A that has a knot and for every $\Gamma \Rightarrow \Delta$ for a basis $\Gamma$ in A, the root and branch of $\Gamma$ are classified by the following four cases: 1) If a new basis in $\Delta$ is formed by the root $\delta \subset \Delta$ and a branch in A but not in $\Gamma$, we call $\delta$ **semi-root**. 2) Similarly, if a new basis in $\Delta$ is formed by the branch $\delta \subset \Delta$, we call $\delta$ **semi-branch**. 3) **pre-discharging tensor links**, if a branch or root $\delta \subset \Delta$ is disappeared from the graph by discharging. 4) A **sub-proved trees**, an other root or branch.*

**Definition 3.** *It is said that $K \Rightarrow K'$ if the semi-root or semi-branch of a basis $K$ grows another basis $K'$. If there exists (multiple or singular) transformations $K \Rightarrow ... \Rightarrow K'$ such that $K'$ has pre-discharging tensor links $\delta$, for every root or branch $\rho$ appeared in the sequence, it is said that **the links $\delta$ depends on $\rho$**.*

**Lemma 1.** *If a transformation on the P-graph A decreases the number of knots in A, the transformation is by the rules $(G_2)$ or $(G_4)$.*

*Proof.* From Figure 6 and 7, the knots are not decreased by the discharging rules. In addition, from Figure 8 and 9, the knots are not decreased by the rules $(G_1)$ and $(G_3)$. Therefore, a transformation that decreases the number of knots is only by the rules $(G_2)$ and $(G_4)$.

**Corollary 1.** *The maximum number of knots decreased by one transformation is only one.*

*Proof.* From Lemma and the transformations by $(G_2)$ and $(G_4)$ shown in Figure 8 and 9.

We characterize the planarity on the proof nets for **LG**. Due to the shapes of links, it can be assumed that an arbitrary knot consists of two contesor links. From Lemma 1, we can assume that if there is a knot $k$ of contesor links $l_1$ and $l_2$ on P-graph $A_1$, an arbitrary sequence of transformations of P-graphs $A_1, A_2, ..., A_n$ for the proof of $A_1$ contains two P-graphs $A_k, A_{k+1}$ obtained by applying a Grishin rule $(G_2)$ or $(G_4)$. Let $K$ be the basis transformed in $A_k \Rightarrow A_{k+1}$. If we restrict the Grishin rules such that only one of $l_1$ and $l_2$ can be discharged (i.e., can depend on the basis $K$), there is no knot on such a system. On the other hand, if there are two links such that each of them depends on the basis $K$ transformed in $A_k \Rightarrow A_{k+1}$, one knot on $A_K$ is vanished in $A_{k+1}$. The transformations of $\mathbf{HLG_\bullet}$ shown in §3.5 satisfies with this condition. Therefore, Theorem 1 holds for $\mathbf{HLG_\bullet}$, and by §3.6, holds for **HLG**.

**Corollary 2.** *The languages of **HLG** exactly corresponds to the tree-adjoining languages and thus is in mildly context-sensitive languages family.*

*Proof.* It is known that the fragment of **LG** that satisfies with planarity can be reduced to Hyperedge Replacement Grammar of rank two ($\mathbf{HR}_2$, often called Hyperedge Replacement Grammar) [13, 14, 8]. Since $HR_2$ is weakly equivalent to lexicalized tree-adjoining grammars, it is proved that the set of languages **HLG** generate on the Lambek system is equivalent to the tree-adjoining languages.

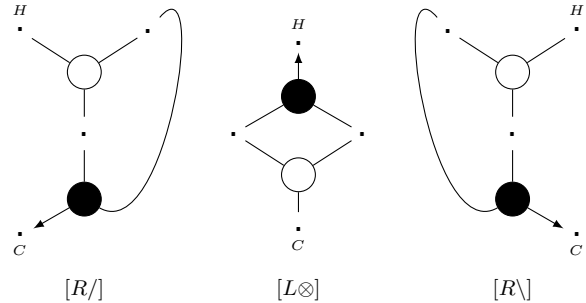

**Fig. 6.** Contractions — Lambek connectives [11]

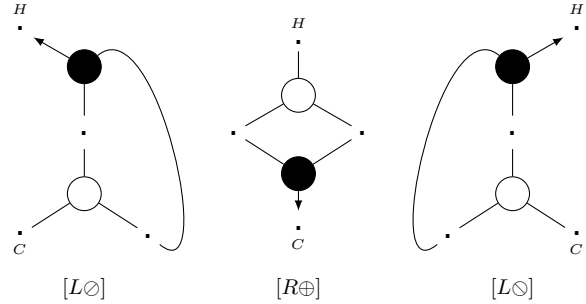

**Fig. 7.** Contractions — Grishin connectives [11]

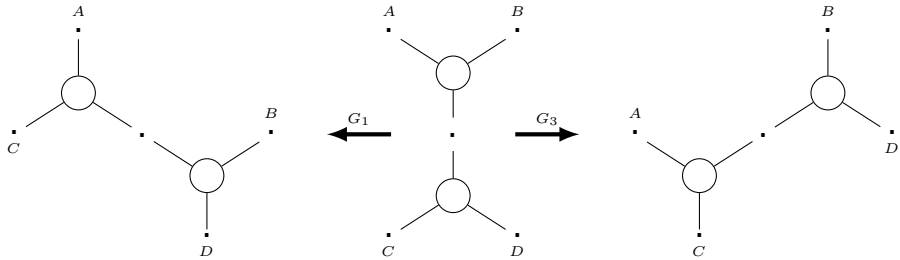

**Fig. 8.** Grishin interactions I — "mixed associativity" [11]

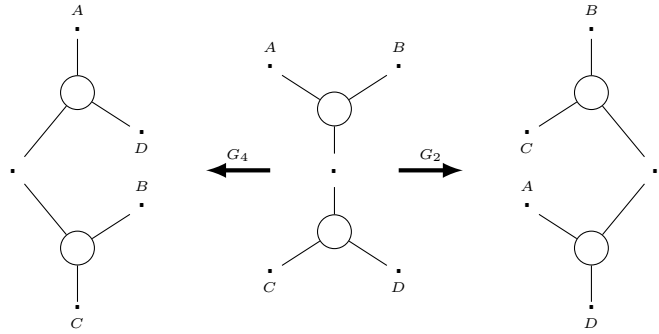

**Fig. 9.** Grishin interactions II — "mixed commutativity" [11]